%% file: main.tex
\documentclass{article} %
\usepackage{iclr2023_conference,times}

\input{math_commands.tex}

\usepackage{hyperref}
\usepackage{url}
\usepackage{enumitem}
\usepackage{tabularx}
\usepackage{booktabs}
\usepackage{array}
\usepackage{lipsum}
\usepackage{wrapfig}
\usepackage{multirow}
\usepackage{graphicx}
\usepackage{scalerel,xparse}
\NewDocumentCommand\emojione{}{\scalerel*{\includegraphics{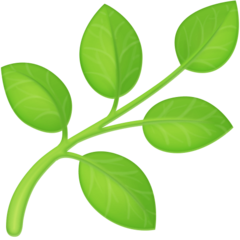}}{X}}

\makeatletter
\def\blfootnote{\gdef\@thefnmark{}\@footnotetext}
\makeatother

\title{MinT\,\emojione{}:\,\,Boosting Generalization in Mathematical Reasoning via \textbf{M}ulti-v\textbf{i}ew Fi\textbf{n}e-\textbf{t}uning}

\author{Zhenwen Liang$^{1*}$, Dian Yu$^{2}$, Xiaoman Pan$^{2}$, Wenlin Yao$^{2}$, Qingkai Zeng$^{1*}$, \\ \bf Xiangliang Zhang$^{1}$, Dong Yu$^{2}$\\
$^1$University of Notre Dame \,\,\,\, $^2$Tencent AI Lab, Bellevue \\
\texttt{\{zliang6,qzeng,xzhang33\}@nd.edu} \\
\texttt{\{yudian,xiaomanpan,wenlinyao,dyu\}@global.tencent.com} \\
}

\newcommand{\cc}{\text{CoT$_\text{clean}$}}
\newcommand{\nc}{\text{CoT$_\text{noisy}$}}
\newcommand{\tr}{\textsc{Tree}}
\newcommand{\eg}{{e.g.}}
\newcommand{\ie}{{i.e.}}

\iclrfinalcopy %
\begin{document}

\maketitle

\begin{abstract}
Reasoning in mathematical domains remains a significant challenge for relatively small language models (LMs). Many current methods focus on specializing LMs in mathematical reasoning and rely heavily on knowledge distillation from powerful but inefficient large LMs (LLMs). In this work, we explore a new direction that avoids over-reliance on LLM teachers, introducing a multi-view fine-tuning method that efficiently exploits existing mathematical problem datasets with diverse annotation styles. Our approach uniquely considers the various annotation formats as different ``views'' and leverages them in training the model. 
By postpending distinct instructions to input questions, models can learn to generate solutions in diverse formats in a flexible manner.
Experimental results show that our strategy enables a LLaMA-7B model to outperform prior approaches that utilize knowledge distillation, as well as carefully established baselines. Additionally, the proposed method grants the models promising generalization ability across various views and datasets, and the capability to learn from inaccurate or incomplete noisy data. We hope our multi-view training paradigm could inspire future studies in other machine reasoning domains.
\end{abstract}

\section{Introduction}
\blfootnote{* This work was done when Zhenwen and Qingkai were interns at the Tencent AI Lab, Bellevue. Codes and data will be released upon paper acceptance.}

Mathematical reasoning, a central aspect of human cognition, has been the subject of inquiry across various disciplines, including philosophy, mathematics, and cognitive science. This capacity, characterized by the analysis of symbolic patterns and logical relationships, and the derivation of conclusions from evidence, is vital in numerous practical applications, such as intelligent education systems \citep{tack2022ai}. The recent development of Large Language Models (LLMs)~\citep{brown2020language,ouyang2022training,touvron2023llama} introduces both a novel challenge and an exciting opportunity for deep learning models to engage in mathematical reasoning.

A significant advancement in machine mathematical reasoning is the discovery of scratchpads~\cite{nye2021work} and chain-of-thought (CoT) reasoning \citep{wei2022chain,kojima2022large} capabilities in LLMs, leading to marked improvements in the accuracy of automated math problem solving. However, this strong mathematical reasoning ability seems to become noticeable only at an extremely large scale, typically exceeding $100$ billion parameters \citep{wei2022emergent} in LLMs. Supporting evidence can be also found in \citep{touvron2023llama}, which reveals that LMs with fewer than $10$ billion parameters struggle to achieve accuracy rates over $20$\% on the GSM8K dataset~\citep{cobbe2021training}, which is essentially comprised of elementary-level math word problems.

To obtain both efficient and effective mathematical reasoning models, one possible approach is to specialize general-purpose LMs in mathematics \citep{fu2023specializing} by distilling the knowledge and abilities from larger teacher models into smaller student models \citep{ho2022large,shridhar2022distilling,magister2022teaching,hsieh2023distilling,liang2023let}. However, this kind of approached faces certain limitations. Firstly, it heavily relies on CoT explanations or additional training samples generated by the larger models to train the smaller student model, and the most common choices for teachers are the GPT series and PaLM-540B, which are inefficient and costly. Secondly, in cases where the larger model might make errors or fail to effectively explain reasoning steps, this could directly impact the quality of the generated training data and subsequently, the learning performance of the student models.

To mitigate the above limitations, instead of relying solely on very large, inefficient LLMs for generating CoT annotations or additional training samples, we focus on an under-explored question: 
\begin{center}
\textit{\textbf{Can we efficiently utilize publicly accessible datasets to develop small LMs specialized in mathematical problem solving?}}
\end{center}
Using existing and annotated datasets can lower manual effort and computational costs compared to having LLMs generate additional annotated data. However, there are also several challenges posed by this direction. Firstly, existing datasets vary significantly in their annotation formats. For instance, the GSM8K \citep{cobbe2021training} dataset presents its solutions in a narrative format detailing step-by-step rationales, while the MathQA \citep{amini2019mathqa} dataset uses flattened programs for annotations, and Ape210K \citep{zhao2020Ape210k}, the largest math word problem dataset, is annotated by equation-based solutions. %
Additionally, when we collect more data from various data sources such as websites, the risk of encountering irrelevant or even incorrect data cannot be disregarded. These differences in annotation style and quality make it difficult to effectively utilize the datasets for training math reasoning models. Empirically, we find that simply merging multiple datasets of different annotation formats cannot always elevate model performance, and in many cases, it even has a negative effect.

To address the above challenges, we propose a \textbf{M}ulti-V\textbf{i}ew Fi\textbf{n}e-\textbf{T}uning (MinT) paradigm. In this context, the disparate annotation methods employed across different datasets are conceptualized as unique ``views'' of mathematical problem solutions. To enhance a model's reasoning ability in a data-efficient way, we not only utilize the original views, but also expand the solution views in existing math word problem datasets by view transformation. Then we concatenate view-specific instructions to the input questions to guide the models to generate solutions in the desired view. We assume that training the model to comprehend various solution views equates to learning different methods of mathematical reasoning, which inherently helps strengthen its reasoning performance and broaden its generalization capabilities. Extensive experimental results support the efficacy of our proposed method, indicating that it fosters a variety of generalizations that contribute to enhancing overall performance across all views. Notably, our paradigm can also be used to incorporate noisy datasets, by regarding them as a new view, to further improve the performance of existing views.
Our contributions can be summarized as follows:

\begin{itemize}[leftmargin=*, itemsep=0pt, topsep=0pt, partopsep=0pt]
\item We propose a multi-view training approach to fine-tune a relatively small language model in the domain of mathematical reasoning. Our approach achieves state-of-the-art performance on four benchmarks, surpassing all baselines. Importantly, this is achieved without the use of large and inefficient teacher models.
\item Our multi-view training method successfully utilizes a large amount of data from both the research community and the broader internet, to enhance the mathematical reasoning capabilities of LMs, which confirms great flexibility and generalizability by effectively handling and learning from data in diverse formats and from various sources.
\item Our extensive experiments demonstrate that our approach performs effectively not only on an externally held-out dataset but also across different LM architectures. This insight demonstrates this approach can be applied more widely, and its potential to inspire future studies aimed at more diverse tasks and backbone models.
\end{itemize}

\input{iclr2023/related_work}

\section{Our Approach}

\subsection{Our Views}
Our method utilizes multi-view training where we conceptualize different annotation styles across datasets as distinct ``views'' of mathematical problem solutions. These views embody a rich collection of solution formats expressed in different levels of symbolism, each with unique nuances and strengths. We categorize the views as follows and show examples for the first three views in Table~\ref{tab:view_example}.

\paragraph{Clean Chain-of-Thought Explanations (\cc)}
The first view, \textbf{clean} \textbf{c}hain-\textbf{o}f-\textbf{t}hought explanations (\cc), is featured in the GSM8K dataset. This annotation style entails a thorough, step-by-step explanation of the solution process. Each intermediary step is clearly elaborated until the final solution is derived. These explanations serve as a detailed guide, illustrate the logical reasoning behind each step and contribute to the comprehension of the entire solving process.

\paragraph{Equation Solutions (EQN)}
The second view, \textbf{eq}uatio\textbf{n} solutions (EQN), presents each question's solution as an equation assembled from a series of operators and quantities, without any explanations. Although this view lacks the detailed explanation provided by CoT solutions, it offers a high-level representation of the solution and is one of the most prevalent annotation formats in datasets such as Ape210K, MathQA, and CM17K. It captures the essence of the problem-solving process in the form of a mathematical expression, making it an efficient and effective format to solve certain types of problems.

\paragraph{Solution Tree Pre-order Traversal (\tr)}
The third view, solution \textbf{tree} pre-order traversal (\tr), is an abstract representation of the solution. Widely adopted by math word problem solvers as suggested in previous studies~\citep{zhang2020graph,liang2022mwp,jie2022learning}, it adopts the pre-order traversal of the solution tree, which avoids the use of parentheses and thus further simplifies the solution grammar compared with EQN solutions. More importantly, this form reflects a goal-driven solving strategy aligned with human reasoning~\citep{xie2019goal}. The expression of solutions in this abstract form fosters efficient solution processing and inference.

\paragraph{Noisy Chain-of-Thought Explanations (\nc)}
The fourth view, \textbf{noisy} chain-of-thought explanations (\nc), is similar to \cc, albeit with noise introduced. This noise may come from incomplete explanations, minor calculation errors, irrelevant domains, or misinterpretation of the problem. This view represents a general category of irrelevant or inaccurate solutions, thereby cannot be used for evaluation and we do not provide examples in Table \ref{tab:view_example}. While challenging, this view reflects the uncertainty and ambiguity in real-world data, providing an opportunity to make models more robust and flexible to different data sources.

\begin{table}[ht]
\centering
\begin{tabular}{lp{10cm}}
\toprule
\multicolumn{2}{p{12cm}}{\textbf{Question}: Beth bakes 4, 2 dozen batches of cookies in a week. If these cookies are shared amongst 16 people equally, how many cookies does each person consume?} \\
\midrule
\bf View & \bf Solution \\
\midrule

\cc    &  Beth bakes 4, 2 dozen batches of cookies for a total of 4*2 = $\ll$4*2=8$\gg$8 dozen cookies. There are 12 cookies in a dozen and she makes 8 dozen cookies for a total of 12*8 = $\ll$12*8=96$\gg$96 cookies. She splits the 96 cookies equally amongst 16 people so they each eat 96/16 = $\ll$96/16=6$\gg$ 6 cookies.      
\\ \addlinespace[2pt]

EQN      & $x$ = 12*(4*2)/16     
\\ \addlinespace[2pt]

\tr     & / * 12 * 4 2 16       \\
\bottomrule
\end{tabular}
\caption{Examples of three views of mathematical solutions.}
\label{tab:view_example}
\end{table}

\subsection{View Transformation}

While the \cc, EQN, and \nc~are provided directly from the original datasets, the third view, \tr, can be derived from the second view EQN, through well-defined algorithms such as~\citep{wang2018translating}. In EQN notation, operators are written between the operands, \eg, ``$A$ $+$ $B$''. \tr notation, also known as Polish notation \citep{lukasiewicz1929znaczeniu}, places operators before the operands, \eg, ``$+$ $A$ $B$'', as shown in Table \ref{tab:view_example}. Besides the transformation between EQN and~\tr, we can also extract all the equations from the \cc~view and then combine and transform them into the EQN view.

\subsection{Multi-View Fine-Tuning}

\begin{figure}
  \centering
   \includegraphics[width=1\textwidth]{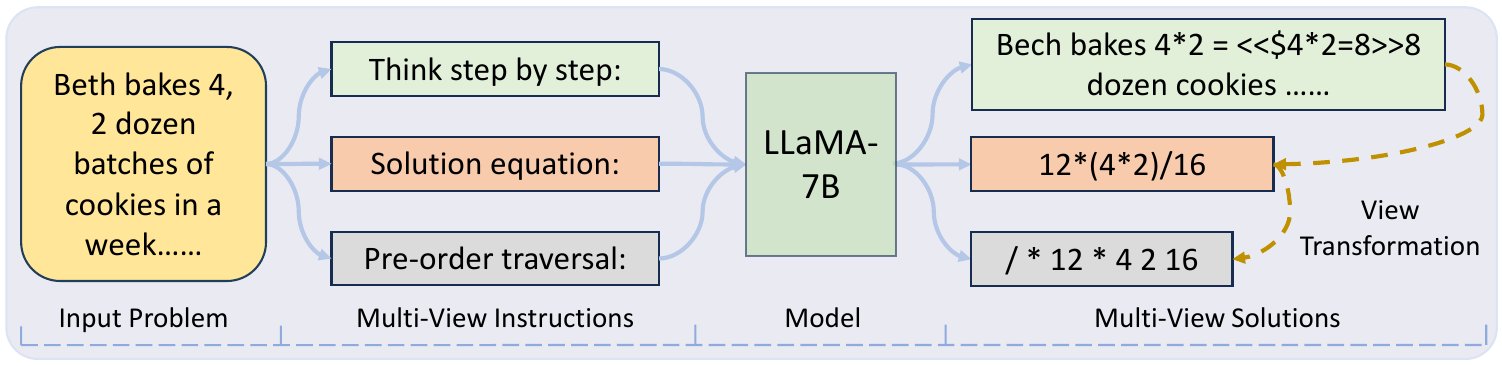}
  \caption{We use MinT to fine-tune a LLaMA-7B model that specializes in math problem solving. First, the original annotation is transformed into multiple different views. Then, the model is trained by instructions to generate different solution forms for one problem.}
  \label{fig:model}
\end{figure}

Our approach leverages a method we refer to as \textbf{M}ulti-V\textbf{i}ew Fi\textbf{n}e-\textbf{T}uning (MinT \emojione{}), which permits us to guide the model in generating different views of solutions by postpending specific instruction strings to the input questions. This results in multiple unique concatenated instructions for each problem, each guiding the model to produce a corresponding view of the answer, as shown in Figure~\ref{fig:model}.

Formally, each question $Q$ is paired with an instruction string $p_i$ drawn from the set $\mathcal{P}$, which includes all possible instructions. When $Q$ is concatenated with $p_i$, it results in a unique string for each question. This provides the necessary guidance for the model to generate a corresponding answer $a_i$ from the answer set $\mathcal{A}$. As such, for each question, we formulate multiple sequences $s_i = Q + p_i + a_i$. Consequently, during the training phase, the model processes a large number of these sequences $s_i$, enhancing its understanding and generalization across multiple views.

To optimize the model, the next-word-prediction loss $L$ is calculated for each sequence:
\begin{equation}
L(s_i) = - \sum_{j=1}^{len(s_i) - 1} \log P(s_{i,j+1} | s_{i,1:j}; \Theta),
\end{equation}
where $P$ denotes the model's conditional probability distribution over the next token, facilitated by the Softmax function of the model's logits, $s_{i,j}$ represents the $j^{th}$ token of sequence $s_i$, and $\Theta$ embodies the model parameters. However, to enhance the model's focus on generating accurate answers, we exclusively backpropagate the loss calculated on the answer part, denoted as $L_{a_i}$:
\begin{equation}
L_{a_i}(s_i) = - \sum_{j=len(Q+p_i)+1}^{len(s_i) - 1} \log P(s_{i,j+1} | s_{i,1:j}; \Theta).
\end{equation}
It ensures that the model focuses on learning to produce precise answers, contributing to its mathematical reasoning ability. During the evaluation, we adopt the same instruction concatenation and assess the model's performance on each individual view. 

To sum up, our approach integrates multiple datasets, each following its own annotation format or ``view''. Our strategy of view transformation serves as an efficient data-utilization method to enhance reasoning generalization, which is achieved by converting data, originally annotated in one view, into multiple diverse views, thus maximizing the utilization of available data. Then, by learning from different views, our model can comprehensively understand mathematical problems, thus improving its reasoning and generalization capabilities. Another advantage of our method is its scalability. MinT can easily accommodate additional views or more data within a single view, thus continually enhancing the model's learning and performance as new data becomes available. Our MinT also has the potential to be integrated with existing knowledge distillation-based methods \citep{ho2022large,shridhar2022distilling,magister2022teaching,hsieh2023distilling,liang2023let}. In this scenario, the output from larger ``teacher'' models can be considered as an additional view. Our model, serving as the ``student'', can then learn to imitate the reasoning processes and outcomes of the teacher model, thus effectively expanding its own problem solving capability.

\section{Experiments}

\subsection{Mathematical Datasets for Training and Testing}

\paragraph{GSM8k}
The GSM8K dataset \cite{cobbe2021training} is a curated set of 8.5K high-quality elementary-level math word problems in English, authored by human problem writers. It is split into approximately 7.5K problems for training and 1K for testing purposes. The problems are annotated with their comprehensive step-by-step solutions, providing the Clean Chain-of-Thought Explanations (\cc) view.

\paragraph{MathQA}
MathQA \citep{amini2019mathqa} contains English mathematical problems from GRE examinations. Nevertheless, some of the problems in this dataset have quality concerns. Several efforts \citep{tan2021investigating,li2021seeking,liang2022mwp} have been conducted to cleanse and filter the MathQA dataset. In our experiment, we adopt the version referenced in \cite{liang2022mwp}, wherein all solutions are re-annotated by an equation composed of the four arithmetic operators and numbers, reflecting the Equation Solutions (EQN) view and we also transform that to the \tr~view.

\paragraph{Ape210k}
The Ape210k dataset \cite{zhao2020Ape210k} is a large-scale, template-rich collection of math word problems (MWPs) in Chinese, containing 210,488 problems and 56,532 solution templates. The view of the solutions in Ape210k mirrors that in MathQA. Our experiment incorporates its 200K training problems and 50K testing problems.

\paragraph{CM17k}
The CM17K dataset \citep{qin2021neural} comprises four types of Chinese MWPs (arithmetic, one-unknown linear, one-unknown non-linear, equation set), which is different from MathQA and Ape210k. Therefore, we only have the EQN view for the solutions in this dataset.

\subsection{Additional Noisy Datasets for Training}

\paragraph{ASDiv-CoT}
The ASDiv dataset \citep{miao2020diverse} consists of 2,305 English MWPs that are diverse in language patterns and problem types. We employ the few-shot CoT predictions of GPT-3 provided by \cite{wei2022chain}\footnote{\url{https://github.com/jasonwei20/chain-of-thought-prompting}.} on this dataset as one of the \nc views for training. With an accuracy of 71.3\%, approximately 30\% of the predictions are spurious. The inclusion of this dataset shows the adaptability and broad applicability of our method to inaccurate LLM-generated data.

\paragraph{ExamQA}
The ExamQA dataset \citep{yu2021self} is a comprehensive Chinese dataset of real-world exams, containing 638k multiple-choice instances across various subjects (e.g., sociology, education, and psychology). We manually filter a subset with 20k problems that contain numbers and equations in their answers by hand-crafted rules. Despite each problem in this subset being annotated with its ground truth and step-by-step solutions, we inevitably introduce many problems that are less relevant to the math subject. This dataset also serves as one of the \nc~views, also showing the generalizability of our approach.

\subsection{Implementation Details}
We employ LLaMA-7B \cite{touvron2023llama} as our backbone %
and perform fine-tuning on the full model. We use Pytorch with DeepSpeed Library to implement the code and use 8 NVIDIA V100 GPUs with 32GB of memory to train our model. We have incorporated a few techniques to ease the computational burden. First, we apply parameter offloading and optimizer offloading and utilize gradient checkpointing to reduce the memory footprint. Additionally, we employ gradient accumulation, effectively enlarging the batch size without demanding additional GPU memory. Lastly, the parameters' precision is set to float-16 (FP16), reducing memory usage and computational requirements. The fine-tuning process lasts for three epochs with a batch size of 64 and a learning rate of 0.00002. Though we find that further fine-tuning could bring slight performance improvements, we limit the number of epochs to three due to efficiency concerns. As a result, the total training time, even while using all the 6 training sets, does not exceed 48 hours.

\subsection{Results}
\subsubsection{Generalization Across Different Views on One Dataset}
Firstly, we investigate the effect of our multi-view fine-tuning on one specific dataset. The GSM8k dataset is selected for this investigation, which is annotated by the 
\cc \, view, which can be conveniently transformed into EQN and \tr~views, thus offering a suitable platform for our investigation. As a baseline, we consider the performance of models that have been fine-tuned on GSM8k using each of the three views individually. Then, we introduce a model that has been fine-tuned with our proposed approach and evaluate it on different views by postpending different instructions.

Table \ref{tab:results_1} shows that augmenting the data with additional views improves performance on all views. Fine-tuning on the original \cc~annotations achieves 35.4\% accuracy on the \cc~test set.
\begin{wraptable}{r}{7cm}
\small
\centering
\begin{tabular}{|c|c|c|}
\hline
\multicolumn{2}{|c|}{\textbf{Prior Work}} & \textbf{Accuracy} \\
\hline
\multicolumn{2}{|c|}{\cite{shridhar2022distilling} (GPT-6B)} & 21.0\% \\
\multicolumn{2}{|c|}{\cite{fu2023specializing} (FlanT5-11B)} & 27.1\% \\
\multicolumn{2}{|c|}{\cite{magister2022teaching} (T5-11B)} & 38.2\% \\
\hline
\textbf{Train View} & \textbf{Test View} & \textbf{Accuracy} \\
\hline
\cc & \cc & 35.4\% \\
EQN & EQN & 30.5\% \\
\tr & \tr & 32.3\% \\
\cc+EQN & \cc & 35.9\% \\
\cc+EQN & EQN & 36.2\% \\
\cc+EQN+\tr & \cc & 36.5\% \\
\cc+EQN+\tr & EQN & 36.9\% \\
\cc+EQN+\tr & \tr & \textbf{37.8\%} \\
\hline
\end{tabular}
\caption{Results on different views of GSM8k.}
\label{tab:results_1}
\end{wraptable}
 Adding EQN and Tree views during training boosts \cc~accuracy to 36.5\%, a 1.1\% absolute improvement. More substantial gains are observed on the Tree view (from 32.3\% to 37.8\%). We can also notice that the performance on \tr~view is the best, where the potential reason is that this view has the simplest grammar, hence it is the easiest view for the model to learn its pattern. Also, we note that prior work utilizes additional high-quality data generated by LLMs and different model architectures. As such, we would like to clarify that our work is orthogonal to previous results and thus, we are not intending to make comparisons, though including them as reference points may be informative.

\subsubsection{Generalization Across Different Datasets with Different Views}

For this investigation, we utilize four different datasets: GSM8k, MathQA, CM17k, and Ape210k. Our baseline for comparison involves prior best results, the single dataset fine-tuning on LLaMA-7B, and simply merging all four datasets for fine-tuning on LLaMA-7B, as shown in Table \ref{table:results_2}. The results show that straightforward merging cannot bring any improvements. Contrastively, it even has a negative effect. Alternatively, with our multi-view learning approach to these four datasets, the model obtains a general improvement across all views when additional training data is added. Another notable finding is that the performance of the \cc~view on the GSM8k dataset gets improved by multi-view fine-tuning with the other three datasets, even the additional datasets actually do not provide any supplementary data in the \cc~view. This outcome shows a promising generalization ability, illustrating the effectiveness of MinT in better leveraging diverse datasets. 
\begin{table}[ht]
\small
\renewcommand\arraystretch{1.05}
\setlength\tabcolsep{5pt}
\centering
\begin{tabular}{c|ccc|cc|c|cc}
\hline
& \multicolumn{3}{c|}{GSM8k} & \multicolumn{2}{c|}{MathQA} & \multicolumn{1}{c|}{CM17k} & \multicolumn{2}{c}{Ape210k} \\ 
\cline{2-9}
\hline
\multicolumn{9}{c}{Single Dataset Baselines} \\
\hline
Prior Best & \multicolumn{3}{c|}{38.2\textsuperscript{a}} & \multicolumn{2}{c|}{76.6\textsuperscript{b}} & \multicolumn{1}{c|}{54.1\textsuperscript{c}} & \multicolumn{2}{c}{70.2\textsuperscript{d}} \\
Single Dataset & \multicolumn{3}{c|}{35.4} & \multicolumn{2}{c|}{79.9} & \multicolumn{1}{c|}{70.1} & \multicolumn{2}{c}{74.0} \\
\hline
\multicolumn{9}{c}{Simple Dataset Mixture} \\
\hline
GSM8k+MathQA & \multicolumn{3}{c|}{36.7} & \multicolumn{2}{c|}{79.7} & \multicolumn{1}{c|}{-} & \multicolumn{2}{c}{-} \\
Ape210k+CM17k & \multicolumn{3}{c|}{-} & \multicolumn{2}{c|}{-} & \multicolumn{1}{c|}{76.0} & \multicolumn{2}{c}{74.9} \\
All Four Datasets & \multicolumn{3}{c|}{35.3} & \multicolumn{2}{c|}{81.0} & \multicolumn{1}{c|}{68.9} & \multicolumn{2}{c}{74.1} \\
\hline
\multicolumn{9}{c}{\textbf{M}ulti-V\textbf{i}ew Fi\textbf{n}e-\textbf{T}uning (MinT \emojione{})} \\
\hline
& \cc & EQN & \tr & EQN & \tr & EQN & EQN & \tr \\
\hline
GSM8k+MathQA & 36.8 & 35.8 & 38.1 & 79.7 & 80.5 & - & - & - \\
Ape210k+CM17k & - & - & - & - & - & 77.1 & 75.9 & 74.3 \\
All Four Datasets & 38.8 & 39.2 & \textbf{40.8} & 81.0 & \textbf{81.3} & \textbf{77.6} & \textbf{76.0} & 74.3 \\
\hline
\end{tabular}
\caption{Experimental results showing the performance of LLaMA-7B with different fine-tuning methods across four datasets. For our method MinT, we report the performance on all available views. The first column shows the training datasets that are used. Prior best results are: a: \cite{magister2022teaching}, b: \cite{liang2022mwp}, c: \cite{qin2021neural}, d: \cite{zhao2020Ape210k}. Some prior best models are based on RNNs, hence not as good as single dataset fine-tuning on LLaMA-7B.}
\label{table:results_2}
\end{table}
\vspace{-3pt}

Furthermore, we observed that the accuracy improvement between the single dataset baseline and our method is more obvious on the GSM8k and CM17k in comparison to the MathQA and Ape210k. A possible explanation could be that MathQA and Ape210k already contain a substantial number of training problems, thereby enabling the learning of problem patterns and solving skills directly from their training sets. Consequently, the contribution of external datasets may not be significant in this case. However, for the more challenging GSM8k and CM17k datasets, our multi-view training could enhance accuracies more effectively. Furthermore, it can be observed that the EQN view performs optimally on the Ape210k dataset, which is different from GSM8k. This could potentially be attributed to the fact that the solutions in Ape210k comprise fewer steps, resulting in relatively simpler equations compared to those in GSM8k and MathQA. Consequently, converting these equations into tree traversals may not substantially simplify the solutions, thereby not improving the model performance. The above two behavior patterns are also echoed in Table \ref{table:results_3}.

\subsubsection{Generalization on \nc~View}
\label{RQ3}
In order to further understand the effects of incorporating external noisy training data, we introduce two additional datasets - ASDiv-CoT and ExamQA. The former provides CoT explanations to problems within the ASDiv dataset, though approximately 30\% of these CoTs are incorrect. The latter, ExamQA, provides CoT to multi-subject exam problems, and while the solutions provided are accurate, a large number of them are less related to mathematical reasoning. Table \ref{table:results_3} presents our experimental results: when we directly add the two noisy datasets for training, there is a slight decrease in accuracy. However, with a specific postfix to differentiate them from the other three views, the overall performance shows an improvement, which demonstrates the potential of using external noisy data to improve the performance on specific downstream tasks. In addition, the results in Table \ref{table:results_2} and \ref{table:results_3} indicate that multilingual data can also complement each other and help improving the general reasoning ability.

\begin{table}[ht]
\small
\renewcommand\arraystretch{1.04}
\setlength\tabcolsep{4.84844pt}
\centering
\begin{tabular}{c|ccc|cc|c|cc}
\hline
& \multicolumn{3}{c|}{GSM8k} & \multicolumn{2}{c|}{MathQA} & \multicolumn{1}{c|}{CM17k} & \multicolumn{2}{c}{Ape210k} \\ 
\hline
\multicolumn{9}{c}{Simple Dataset Mixture} \\
\hline
Four Datasets & \multicolumn{3}{c|}{35.3} & \multicolumn{2}{c|}{81.0} & \multicolumn{1}{c|}{68.9} & \multicolumn{2}{c}{74.1} \\
\hline
Four Datasets + Two Noisy Datasets & \multicolumn{3}{c|}{31.9} & \multicolumn{2}{c|}{79.7} & \multicolumn{1}{c|}{71.3} & \multicolumn{2}{c}{73.2} \\
\hline
\multicolumn{9}{c}{\textbf{M}ulti-V\textbf{i}ew Fi\textbf{n}e-\textbf{T}uning (MinT \emojione{})} \\
\hline
& \cc & EQN & \tr & EQN & \tr & EQN & EQN & \tr \\
\hline
Four Datasets & 38.8 & 39.2 & 40.8 & 81.0 & 81.3 & 77.6 & 76.0 & 74.3 \\
Four Datasets + Two Noisy Datasets & 39.2 & 39.7 & \textbf{42.4} & 82.0 & \textbf{82.3} & \textbf{78.8} & \textbf{77.0} & 76.1 \\
\hline
\end{tabular}
\caption{Experimental results showing the performance of LLaMA-7B with different fine-tuning methods. For our method MinT, we report the performance on all available views. Four datasets indicate the combination of four clean datasets - GSM8k, MathQA, CM17k and Ape210k, while two noisy datasets are ASDiv-CoT and ExamQA.}
\label{table:results_3}
\end{table}
\vspace{-14pt}
\subsubsection{Evaluation on Held-out Dataset}
\begin{wrapfigure}{r}{0.485\textwidth}
  \centering
  \includegraphics[width=0.485\textwidth]{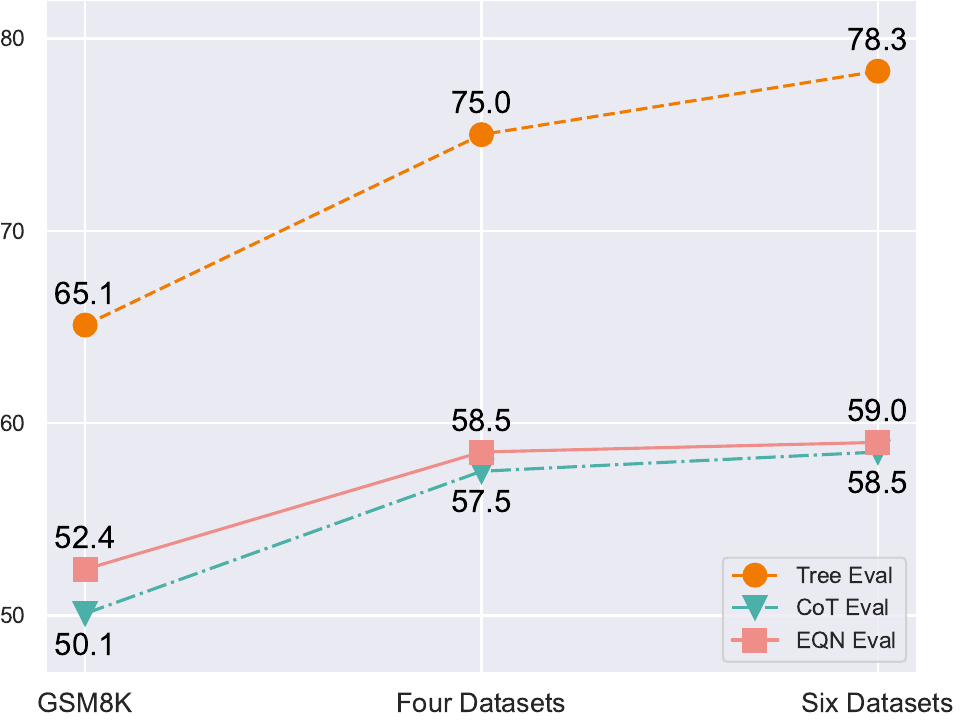}
  \caption{Experimental results on the MAWPS dataset. X-axis indicates the training datasets and Y-axis indicates the accuracy.}
  \label{fig:mawps}
\end{wrapfigure}
In order to further assess the multi-view problem solving abilities of our method, we evaluated it on the held-out dataset, MAWPS \cite{koncel2016mawps}, which contains 2,373 English MWPs annotated with Equation Solutions (ES) view. It integrates several earlier datasets in  \citet{hosseini2014learning}, \citet{kushman2014learning},\cite{koncel2015parsing} and \cite{roy2015solving} and thus serves as a comprehensive benchmark. Three training data settings are used: GSM8k only, four clean datasets, and all six datasets, where respective models are all trained with MinT. 
Our results in \ref{fig:mawps} are similar to the observations from our previous experiments: increasing the number of datasets used in training boosts performance across all views. It is noteworthy that although the MAWPS dataset is originally annotated using the EQN view, our model manages to attain an accuracy of 58.5\% when attempting to solve problems using step-by-step CoTs. This finding indicates that the problem solving ability acquired from the training datasets can indeed be transferred to the held-out datasets. More interestingly, it suggests that our method could serve tasks like multi-view data annotation.

\subsubsection{Adaptivity on Different Backbones}
\begin{wrapfigure}{r}{0.525\textwidth}
  \centering
  \includegraphics[width=0.525\textwidth]{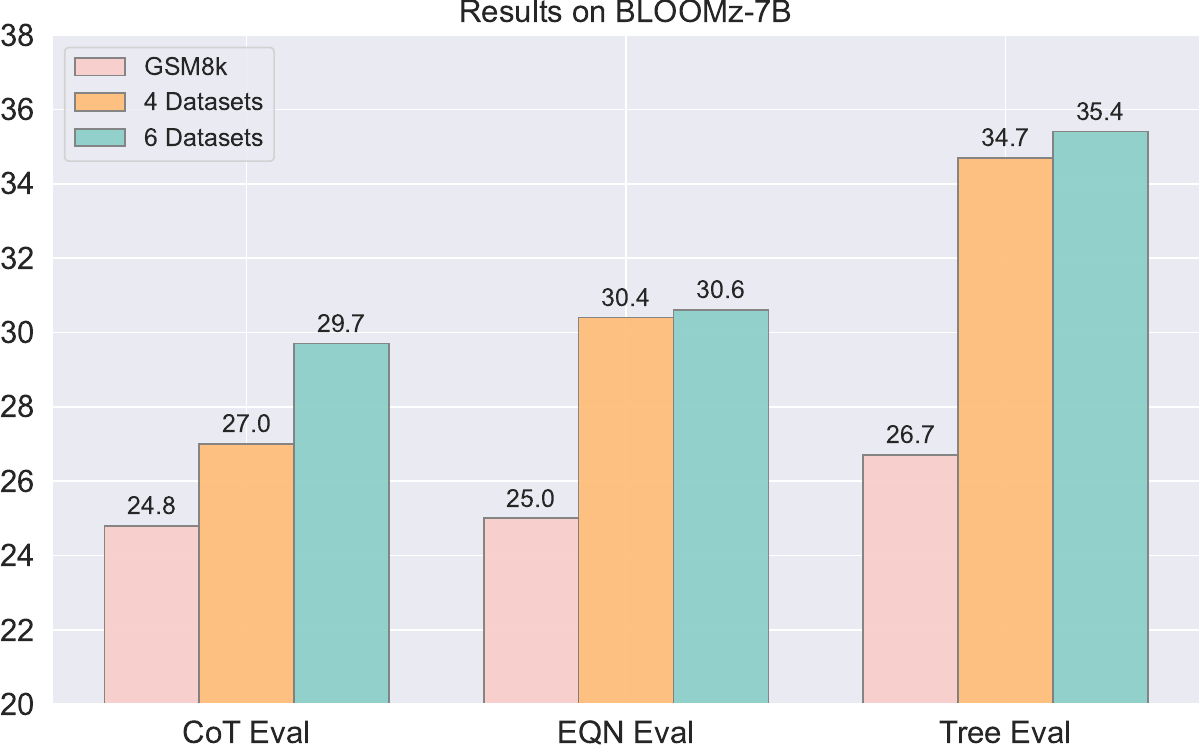}
  \caption{Experimental results with BLOOMz-7B backbone on GSM8K. X-axis indicates the evaluation views and Y-axis indicates the accuracy.}
  \label{fig:bloom}
\end{wrapfigure}
In this experiment, we replace our original backbone model LLaMA-7B with another state-of-the-art model, namely BLOOMz-7B \cite{muennighoff2022crosslingual}. This substitution allows us to observe how well our method adapts to different language model architectures. 
We also employ three models for this task: one trained on the GSM8k dataset, another trained on a combination of four clean math datasets, and a third trained on a total of six datasets. For each of these models, multi-view fine-tuning is employed during the training process. 
As illustrated in Figure \ref{fig:bloom}, our method demonstrates the same pattern on the BLOOMz-7B backbone compared to the LLaMA-7B backbone. With the aid of multi-view training, incorporating additional training data can further enhance performance. This means that the success of our method is not restricted to the LLaMA-7B model; it also extends to other language models. This consistent performance across different backbones validates the robustness of our approach and supports its potential applicability in a wider range.

\section{Discussion}
\subsection{Conclusion}
In this paper, we propose MinT \emojione{}, a novel multi-view fine-tuning approach to enhance the mathematical reasoning capabilities of language models. By framing diverse annotation formats across datasets as distinct ``view'' of solutions, our method enables models to learn from these unique problem-solving perspectives. We also provide a data-efficient view-transformation strategy, expanding the model's ability to generalize and reason by converting the data annotated in one view into multiple different views. Through MinT, our model exhibits strong performance on multiple benchmarks, outperforming prior knowledge distillation-based techniques. Our experiments demonstrate promising generalization ability and adaptivity to noisy data and across model architectures.

\subsection{Broader Impact}
We believe MinT provides a scalable and flexible approach for specialized reasoning, with the potential for broader applicability beyond mathematical domains. Many other reasoning tasks, such as commonsense or symbolic reasoning, can be solved through diverse paths, sometimes with multiple valid solutions. Investigating how to leverage multi-view training for these types of flexible reasoning is an exciting future direction.

Furthermore, the multi-view technique demonstrates effective control over language model generations. By guiding the model with simple instruction strings, we can take advantage of even incomplete and irrelevant data, while still producing high-quality results for downstream tasks. This opens possibilities for combining the benefits of large-scale foundational training and task-specific tuning.

\subsection{Limitations and Future Work}
While we demonstrate generalization on held-out
datasets, evaluation on more diverse views and tasks would further validate the capabilities of our proposed method. Also, there may be optimal combinations and proportions of data across views that our current work does not explore in depth. Another unexplored aspect is whether multiple views can augment each other through shared learning, potentially enhancing the overall accuracy. Techniques such as ensemble learning or majority voting could provide avenues for further improvement in this regard. We leave these possibilities in our future work.

\bibliography{iclr2023_conference}
\bibliographystyle{iclr2023_conference}

\end{document}

%% file: math_commands.tex
\usepackage{amsmath,amsfonts,bm}

\def\eqref#1{equation~\ref{#1}}

\def\1{\bm{1}}

\DeclareMathAlphabet{\mathsfit}{\encodingdefault}{\sfdefault}{m}{sl}
\SetMathAlphabet{\mathsfit}{bold}{\encodingdefault}{\sfdefault}{bx}{n}

%% file: iclr2023/related_work.tex
\section{Related Work}

\subsection{Multi-View Learning}
In traditional machine learning, multi-view learning often refers to semi-supervised co-training algorithms~\citep{nigam2000analyzing,sun2011robust}. These algorithms exploit multiple views of data to iteratively learn separate classifiers, each of which provides predicted labels for the unlabeled data of the others, \ie, semi-supervised setting. Another thread of multi-view learning lies in clustering methods. These methods aim to partition the data across multiple views, which provide complementary information to each other, to obtain a more refined representation of the data \citep{bickel2004multi,li2015large,cao2015diversity}.

More recently, the concept of multi-view learning has been extended to deep learning~\citep{song-2020-structural}. For instance, \cite{wang2022multi} employs multiple transformers to learn a comprehensive embedding for speaker recognition. Similarly, \cite{zhong2023knowledge} utilizes views based on context, syntax, and knowledge to analyze the sentiment of sentences. Moreover, \cite{allen2022towards} shows that the multi-view learning strategy can be associated with ensemble learning and knowledge distillation techniques to improve the accuracy of image classification tasks.

While traditional multi-view learning approaches aim to improve data representation through understanding different perspectives, our work takes a different path by focusing on mathematical problem solving, which is unlike deterministic tasks such as image classification or sentiment analysis. This free-form generation task can be characterized by its diverse solution views shown in Table \ref{tab:view_example}. This situation is further complicated by the existence of multiple acceptable solutions for a problem, as discussed in \cite{hong2021learning}. Given these conditions, our goal is not to prioritize any single view, but rather to enhance the accuracy of solutions across all views. In our context, the term ``generalization'' indicates the collaboration and mutual benefit among all the views, differing from the traditional concept of generalizing to a specific view or deterministic task. We hope this work can broaden the application and impact of multi-view learning techniques.

\subsection{Math Word Problem Solving}
Solving math word problems is a common benchmark for evaluating the mathematical reasoning capabilities of NLP models \cite{amini2019mathqa,patel2021nlp,cobbe2021training}. Earlier approaches rely on statistical and rule-based parsing \cite{hosseini2014learning,koncel2015parsing}, then transition to Seq2Seq-based neural networks \cite{xie2019goal,zhang2020graph,jie2022learning}. Recently, Large Language Models (LLMs) have demonstrated success in solving math word problems, surpassing fine-tuned baselines through Chain of Thought (CoT) prompts \cite{wei2022chain,kojima2022large}.

Another thread of research has also explored the distillation of knowledge from LLMs to smaller models \cite{ho2022large, magister2022teaching, shridhar2022distilling, hsieh2023distilling, liang2023let}. These papers primarily leverage LLMs for generating reasoning steps, segmenting problems, or creating customized exercises to train smaller models.

Our research, however, investigates a different approach: using publicly accessible datasets in an efficient way to train effective, smaller LMs for mathematical problem solving. We thus offer a unique perspective, being orthogonal to previous research efforts.